\pdfoutput=1

\documentclass[11pt]{article}

\usepackage[preprint]{acl}

\usepackage{times}
\usepackage{latexsym}

\usepackage[T1]{fontenc}

\usepackage[utf8]{inputenc}

\usepackage{microtype}

\usepackage{inconsolata}

\usepackage{graphicx}
\usepackage{amsmath}
\usepackage{amssymb}
\usepackage{algorithm}
\usepackage{algorithmic}
\usepackage{booktabs}

\title{Fast Exact Retrieval for Nearest-neighbor Lookup (FERN)}

\author{Richard Zhu \\
  Princeton University \\
  \texttt{ryzhu@princeton.edu} \\}

\begin{document}
\maketitle
\begin{abstract}
Exact nearest neighbor search is a computationally intensive process, and even its simpler sibling --- vector retrieval --- can be computationally complex. This is exacerbated when retrieving vectors which have high-dimension $d$ relative to the number of vectors, $N$, in the database. Exact nearest neighbor retrieval has been generally acknowledged to be a $O(Nd)$ problem with no sub-linear solutions. Attention has instead shifted towards Approximate Nearest-Neighbor (ANN) retrieval techniques, many of which have sub-linear or even logarithmic time complexities. However, if our intuition from binary search problems (e.g. $d=1$ vector retrieval) carries, there ought to be a way to retrieve an organized representation of vectors without brute-forcing our way to a solution. For low dimension (e.g. $d=2$ or $d=3$ cases), \texttt{kd-trees} provide a $O(d\log N)$ algorithm for retrieval. Unfortunately the algorithm deteriorates rapidly to a $O(dN)$ solution at high dimensions (e.g. $k=128$), in practice. We propose a novel algorithm for logarithmic Fast Exact Retrieval for Nearest-neighbor lookup (FERN), inspired by \texttt{kd-trees}. The algorithm achieves $O(d\log N)$ look-up with 100\% recall on 10 million $d=128$ uniformly randomly generated vectors.\footnote{Code available at https://github.com/RichardZhu123/ferns}
\end{abstract}

\section{Introduction}

\label{sec:intro}
Vector retrieval is pervasive, underlying search engines, transformers, and open-book language models. At heart, one of key attributes of computing systems lie in their ability to retrieve knowledge. Sometimes, when this knowledge is sufficiently broad --- and a powerful enough retrieval architecture is built --- these computing systems may even be so good at retrieving relevant knowledge that they appear to reason \cite{bubeck2023sparks}.

Given the power of knowledge retrieval for both commercial and academic pursuits, significant energy has been devoted towards effectively converting various types of data into vectors, from words \cite{mikolov_efficient_2013}, images \cite{radford_learning_2021}, and audio \cite{radford_robust_2022} to documents, sentences, and paragraphs \cite{dai_document_2015}.

In this work, we differentiate between look-up and search. Look-up involves the retrieval of vectors guaranteed to be contained in the database, while search involves the retrieval of queries not necessarily contained in the database. Retrieving queries without exact matches may involve instead retrieving that vector's nearest neighbors. The definition of nearest can be further disambiguated into Euclidean or cosine similarity measures, among others. Note that under the hood, a Euclidean distance-based nearest neighbor algorithm can be easily adapted to be cosine similarity-based simply by dividing each vector in the database by its magnitude during insertion. During look-up, the query vector is then also divided by magnitude. The resulting nearest neighbors we obtain are also such by cosine similarity.

The scalable and effective look-up of large numbers of high dimensional vectors is thus desired. While the vanilla hashmap algorithm provides $O(1)$ time complexity for scalars, extending to $O(d)$ for vectors in d-dimensional space, this holds only when the cardinality of the hash function range is large relative to the number of elements, $N$. When the number of elements becomes large relative to the number of bins, $b$, finding the key within each bin becomes a linear search problem. Since we expect each bin to have $\frac{N}{b}$ collisions, the time complexity for look-up is $O(\frac{Nd}{b})$. While we can also get fast look-up with a heap in low dimensions, FERN can be thought of as an extension of the heap to high dimensions while also presenting a novel approach that could lead to sub-linear vector search. 

\section{Related work}
\label{sec:relwo}
Prior work has taken 3 major directions to attain fast nearest neighbor search. The approaches involve bucketing, divide and conquer, or graph-based approaches. These techniques are specifically tuned to work well for vectors in high dimensional space, which should be unsurprising since each of these techniques is really just an extension of familiar 1-d concepts: hashmaps, binary search, and breadth first search. I believe there are key learnings that can be taken from both exact and approximate retrieval and search settings, even though we are interested in the exact variant. Some new techniques such as Certified Cosine certificates \cite{francis-landau_exact_2019} offer structure-agnostic tweaks to speed up performance.

\subsection{Bucketing}

Locality Sensitive Hashing \cite{indyk_approximate_1998} and k-means \cite{lloyd_least_1982, macqueen_methods_1967}, including more recent variants like k-means++ \cite{arthur_k-means_2007}, are two techniques that use bucketing to group database vectors with their nearby peers. Since it is hard to deal with queries that fall on the boundaries of adjacent clusters, these algorithms are approximate search algorithms. Both of these algorithms in practice take linear time for sufficiently high dimensions and large numbers of elements.

\paragraph{Divide and conquer}
kd-trees provide strong algorithms for pruning, with new variants attempting to decrease constant factors in the time complexity. \cite{zhang_large-scale_2012} A recent work \cite{ram_revisiting_2019} attempts to create partitions based on random rotations of the dataset achieve the same search accuracy guarantees as RPTree \cite{dasgupta_randomized_2013} but with $O(d\log d+\log n)$ time complexity for approximate search.

\paragraph{Graph-based approaches}
Recent graph-based approaches to search, such as Navigable Small World (NSW) graphs \cite{malkov_approximate_2014} and a later variant involving layers of NSWs, provide increasing granularity. NSWs are graphical representations of databases where each pair of nodes is connected by a small number of hops. Further optimizations have been presented \cite{fu_fast_2019, jayaram_subramanya_diskann_2019}.

\section{FERN}
\label{sec:fern}

\paragraph{Goal} We aim to build an algorithm that satisfies two primary goals: we must be able to perform quick look-up on vectors guaranteed to be contained in the database and we must be able to quickly insert vectors. When designing the algorithm, we initially assume our database contains $N$ vectors, each spanning $d$ dimensional space. The $i$-th vector $v_i$ is defined as follows
\begin{align*}
    &\mathbf{v_i} = [v_{i,1}, v_{i,2}, \ldots, v_{i,d}]^\top \\
    &\text{where } v_i \sim \mathcal{R}(-1, 1)
\end{align*}

This process effectively generates vectors lying within a $d$ dimensional "ball"-like shape of radius 1 in each direction. The directions in which the vectors point are also evenly distributed direction-wise. We discover later however, that the algorithm we design with this simplifying assumption maintains logarithmic time lookup for $\mathbf{v_i}$ of any arbitrary direction and length. 

In terms of time complexity, we define quick as anything taking logarithmic time --- this means that lookup in a database of ten billion vectors should only take seven times longer than lookup in a database of a thousand vectors. That is remarkable, because a naive linear search would take ten million times longer, a nearly intractable time difference at scale. Since a logarithmic lookup time and linear space complexity is state of the art (SoTA), we believe a key contribution of our work is an alternative data structure and algorithm that achieves SoTA while simultaneously providing the capacity to be extended to logarithmic-time nearest neighbor search, given its unique approach to dividing the vectors by hyperplanes defined based on the vectors in the database rather than measuring along a specific direction like the traditional kd-trees process. Each node is an object that stores a vector, pointers to the left and right children, and a pointer to that node’s parent node.  This results in a binary tree with undirected edges. While we ultimately implement retrieval using a queue structure, this bidirectional edge only adds marginal complexity to the algorithm and underlying data structure while enabling a backtracking-based traversal method. The queue-based method emulates a level-order traversal of candidate nodes while a stack-based backtracking-based traversal method (that fully explores a specific path before backtracking; exploring each sibling node that could not be pruned without potentially missing the nearest neighbor) emulates a depth-first search.

\paragraph{Methodology} We design a novel algorithm that is a variant of kd-trees, but has the capacity to perform better at higher dimensions. Broadly, the structure is a binary tree. Each node that has both left and right children defines a hyperplane using the vectors of its left and right children as support vectors.

The tree is constructed so that all vectors in each child's subtree are on the same side of the hyperplane as that child. This allows us to perform vector look-up in logarithmic time, provided that vectors are added to the database in a sufficiently random manner. It is feasible, however, for an adversarial insertion process to result in a heavily imbalanced tree and consequently worst-case linear look-up time. While we don't observe this as an issue in practice, we can resolve the issue by implementing a slightly more complicated variant of FERN - using a variant of the Red-Black Tree technique \cite{guibas_dichromatic_1978} to guarantee balanced trees, logarithmic depth, and thus logarithmic retrieval time complexity.

There are two key components to our algorithm: one method for insertion (Algorithm \ref{alg:insert}) and another for lookup (Algorithm \ref{alg:lookup}). 

The insertion algorithm (Algorithm \ref{alg:insert}) is fairly concise. When inserting a vector into the tree, it is placed at the root if the tree has not been initialized yet. Otherwise, if the current node is missing a left or right child, we insert the vector as a child node. If the node is a leaf node --- that is, missing both left and right children --- then the left child is always inserted first, before the right child.

\begin{algorithm}
\caption{FERN Insertion}
\begin{algorithmic}[1]
\STATE \textbf{Function} Insert($\text{vector}$)
    \IF{$\text{root not initialized}$}
        \STATE $\text{root} \gets \text{VectorNode(null)}$
    \ENDIF
    \STATE $\text{node} \gets \text{root}$
    \WHILE{$\text{True}$}
        \IF{$\text{no left child}$}
            \STATE $\text{set left node to vector}$
            \STATE \textbf{break}
        \ELSIF{$\text{no right child}$}
            \STATE $\text{set right node to vector}$
            \STATE \textbf{break}
        \ELSIF{$\text{vector closer to left child}$}
            \STATE $\text{node} \gets \text{node.left}$
        \ELSE
            \STATE $\text{node} \gets \text{node.right}$
        \ENDIF
    \ENDWHILE
\end{algorithmic}
\label{alg:insert}
\end{algorithm}

During insertion, if a node has both left and right children, then we set the current node instead to the child node that is closest to the vector we are inserting. That is, if we form a hyperplane from the set of points equidistant to both left and right children, then we set the current node to the left child if the vector to be inserted lies on the same side of the hyperplane as the left child, otherwise we set the current node to the right child.

The result of this insertion algorithm is that --- for a balanced binary tree --- we get a maximum tree depth of $O(\log N)$ where $N$ is the number of elements in the database. Insertion time per vector is thus $O(\log N)$ since we only visit one node per depth level. 

When looking up vectors from the data structure, we demonstrate a method (Algorithm \ref{alg:lookup}) that has a per-vector retrieval time proportional to the maximum depth of the tree, since we only look at one node per depth level. However, when extended to search settings where the query is not known to be contained in the database, retrieval time becomes proportional to the number of elements in the tree. We can no longer automatically prune any queries that lie close to the hyperplane boundary since there is a possibility that the nearest neighbor and query may lie on different sides of the hyperplane.

Ostensibly, we would expect a non-negligible proportion of vectors to be sufficiently far from the hyperplane to be pruned. However, we notice that in practice, as the dimensionality of the vectors increase, so too does the proportion of vectors lying close to the boundary. This makes sense intuitively since we are trying to project increasingly higher dimensions of vectors onto a 1-d line (the line normal to the hyperplane and passing through both support vectors). During retrieval, we effectively perform the depth-first or level-order search, as described previously. For a balanced tree with strong boundaries (that is, most queries lie far away from the hyperplane),  per-vector time should be logarithmic with respect to the number of elements already present in the database (another word for our proposed data structure) at insertion-time. However, it becomes linear otherwise. We first define \texttt{mip} and \texttt{mip\_vec}, the distance to the nearest neighbor found thus far and the vector representing the nearest vector retrieved thus far. We then create a queue and insert the root node.

\begin{algorithm}[h]
\caption{FERN Lookup}
\begin{algorithmic}[1]
\REQUIRE $\text{query}$
\ENSURE $\text{mip\_vec}$
\STATE $\text{mip}, \text{mip\_vec} \gets \infty, \text{None}$
\STATE $\text{curr} \gets \text{None}$
\STATE $\text{queue} \gets [\text{self.root}]$

\WHILE{$\text{queue not empty}$}
    \STATE $\text{curr} \gets \text{oldest element in queue}$
    \STATE $\text{update mip, mip\_vec if curr closer to query}$
    \IF{$\text{curr has both left and right children}$}
        \IF{$\text{query is closer to left child}$}
            \STATE $\text{queue.append(curr.left)}$
            \IF{$\text{query close to boundary}$}
                \STATE $\text{queue.append(curr.right)}$
            \ENDIF
        \ELSE
            \STATE $\text{queue.append(curr.right)}$
            \IF{$\text{query close to boundary}$}
                \STATE $\text{queue.append(curr.left)}$
            \ENDIF
        \ENDIF
    \ELSIF{$\text{curr has left child only}$}
        \STATE $\text{queue.append(curr.left)}$
    \ELSIF{$\text{curr has right child only}$}
        \STATE $\text{queue.append(curr.right)}$
    \ENDIF
\ENDWHILE

\RETURN $\text{mip\_vec}$
\end{algorithmic}
\label{alg:lookup}
\end{algorithm}

We then continuously pop a node from the head of the queue until the queue is empty. Each time we pop a node, we check whether its vector is closer to the query vector than the current best candidate for nearest neighbor, \texttt{mip\_vec}, which is a Euclidean distance \texttt{mip} away from the query. For lookup, we are looking for an exact match, so we are seeking an \texttt{mip} of 0. If the node has a left child only or a right child only (the latter should never happen, but we have it as a redundancy against exceptions) then we add that node to the queue. Otherwise, if both children exist then we add to the queue the node that shares the same side of the hyperplane as the query. For lookup cases, we consider any query to be sufficiently far from the boundary that only one child node needs to be added to the queue per node. After all, whether a query is close to the boundary is somewhat arbitrary and the exact function definition depends on whether we are performing lookup or search.

For mapping applications, we can add an additional variable, \texttt{data} (a byte array), to the \texttt{Node} class.

In Algorithms \ref{alg:insert} and \ref{alg:lookup}, ``closeness" is quantified by Euler distance, and the ``boundary" is the midpoint between left and right child nodes.

\section{Experimental results}
\label{sec:exp}

During experiments, we typically utilize $d=128$ with the same uniform distribution previously assumed. While this may not be characteristic of all data distributions, we note that our architecture is actually agnostic to the distribution of the vectors being inserted. What matters (in terms of potential effects on performance) is the order in which vectors are inserted based on their relative positions.

To run our experiments, we use the Intel Xeon Platinum 8380 CPU (2.30 GHz), the same processor used for running the popular \texttt{ann-benchmark} \cite{bernhardsson_erikbernann-benchmarks_2024}. For values of $N$ we use $10^4$, $5*10^4$, $10^5$, $5*10^5$, and $10^6$, $5*10^6$, $10^7$. The last setting, equivalent to look-up on 10 million vectors, has comparable values of $N$ and $d$ to many of the Euclidean distance based benchmarks in \texttt{ann-benchmark}. We notice a nearly perfect logarithmic time complexity, and at $N=10^7$ we run approximately 3000 retrievals/second without additional optimizations. 

\section{Discussion}
\label{sec:disc}

During our experiments, we noticed that in the \underline{search} modality, system dynamics can change drastically based on the dimensionality of the vectors. We experiment with different ways of defining the hyperplanes and various algorithms that would balance or repair the tree to try to guarantee logarithmic retrieval for large $d$ and $N$. Unfortunately, while these algorithms almost universally gave logarithmic time complexity for 100\% recall, the performance broke down drastically beyond $d=2$ or $d=3$. In particular, we note the importance of having well-defined hyperplane boundaries.

\paragraph{Boundary sharpness} We want to be able to maximize pruning since we achieve $O(\log_2N)$ time complexity when we prune 50\% of the nodes in the database each time we measure the distance between a node and the query). Indeed, note that since a $k$-means based nearest neighbors search allows us to prune up to $N/k$ nodes per comparison, we might wonder why $2$-means search doesn't achieve $O(\log N)$ complexity. It's because $k$-means has an $O(N/k + k)$ time complexity, the $N/k$ term means that we would need to do a linear number of searches regardless of the number of clusters.

We observe empirically that the proportion of nodes in the database that are visited increases sharply when there are more nodes that are closer to the hyperplane than the support vectors that define the plane. That is to say, during the insertion process, there will be vectors that lie between a support vector and the hyperplane. Now when we're retrieving, the query may lie on one side of the hyperplane but its nearest neighbor may be one of these "in-between" nodes on the other side of the hyperplane. This means we now must be much more prudent when pruning which decreases the proportion of vectors that are pruned and thus increases the time complexity.

\begin{figure}
    \centering
    \includegraphics[width=.5\textwidth]{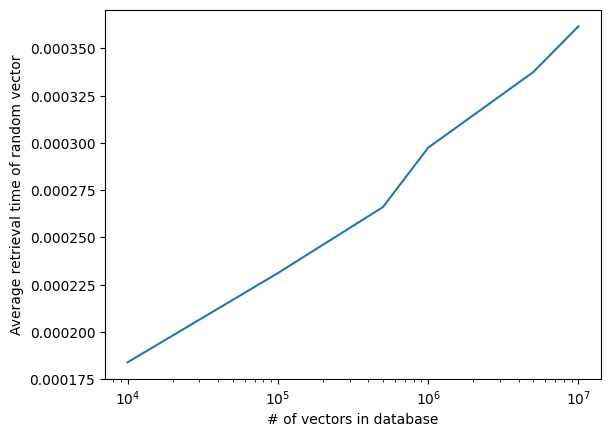}
    \caption{FERN lookup with vectors where $d=128$ and look-up time is averaged over 1000 vectors randomly sampled from the database}
    \label{fig:look-up}
\end{figure}

\begin{figure}
    \centering
    \includegraphics[width=.5\textwidth]{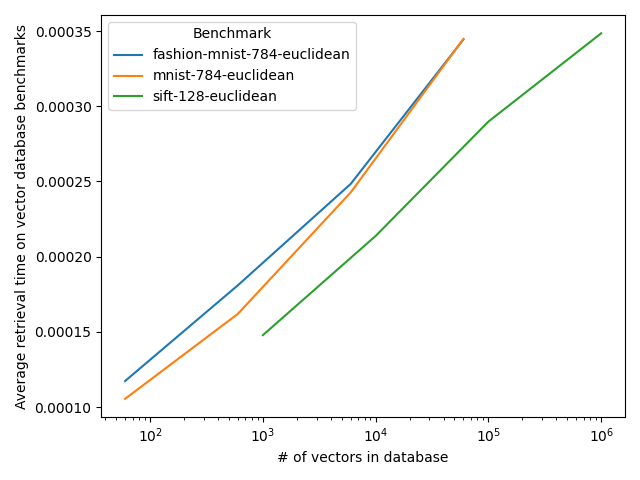}
    \caption{FERN lookup using the train portion (60k-100k vectors) of popular Euclidean-distance-based vector retrieval benchmarks and look-up time is averaged over 1000 vectors randomly sampled from the database. We evaluate 4 decades on each dataset, which is why SIFT-128-Euclidean evaluation starts with vector databases of size $10^3$ rather than $600$}
    \label{fig:ann-bench}
\end{figure}

However in the \underline{look-up} modality, we achieve logarithmic time complexity on both a vector database of dimension $d=128$ and size $10^7$ that are randomly and uniformly generated (Figure \ref{fig:look-up}) and three Euclidean benchmarks (Table \ref{tab:dataset_summary}) from \texttt{ann-benchmark} (Figure \ref{fig:ann-bench}), using the larger \texttt{train} splits ($N=60,000$ to $N=1,000,000$). We observe logarithmic time complexity over a diverse dimensions and vector distributions. 

\begin{table}[ht]
\centering
\small 
\begin{tabular}{@{}lccc@{}} 
\toprule 
Dataset & Dim. & Train Size & Test Size \\ 
\midrule 
Fashion-MNIST & 784 & 60,000 & 10,000 \\
MNIST & 784 & 60,000 & 10,000 \\
SIFT & 128 & 1,000,000 & 10,000 \\
\bottomrule 
\end{tabular}
\caption{Properties of ANN Benchmark datasets used for evaluation}
\label{tab:dataset_summary}
\end{table}

\section{Conclusion}
\label{sec:conc}
We are able to achieve our goal of creating a novel vector database structure that achieves state of the art look-up time complexity that is logarithmic in the number of vectors. The algorithm presented here, FERN, maintains 100\% recall while performing lookup on vectors in high-dimensional space (e.g. $d=128$ to $d=784$) with $N$ varying from $10^2$ to $10^7$, and presents a potential path towards a data structure and algorithm that will allow for the first sub-linear exact nearest neighbor retrieval process.

We believe that the exact process for attempting to perform binary search on a vector database requires carefully defined hyperplanes, which presents an area for further work. We find the "fixing" step of the Red-Black tree algorithm to be particularly inspirational as a direction of future work. Evaluation on additional datasets and further investigations of recall-retrieval-time trade-offs could help in the pursuit of sub-linear search.

We further also believe that an alternative for hyperplanes is to use a graph based approach, similar to the approach taken in many recent works \cite{malkov_approximate_2014, fu_fast_2019, jayaram_subramanya_diskann_2019}, since this could allow us to more easily divide the database in a well-defined and easy to update way. Overall we are excited by the potential and hope to further develop this algorithm in pursuit of sub-linear exact search.

\section*{Acknowledgements}
The author thanks Edo Liberty, Huacheng Yu, Runzhe Yang, Nataly Brukhim, Douglas Downey, and Matthijs Douze for their invaluable advice and feedback during the writing of this paper.


\bibliography{custom}

\appendix

\end{document}